\documentclass[lettersize,journal]{IEEEtran}
\usepackage{color}
\usepackage{amsmath,amsfonts}
\usepackage{algorithmic}
\usepackage{algorithm}
\usepackage{array}
\usepackage[caption=false,font=normalsize,labelfont=sf,textfont=sf]{subfig}
\usepackage{textcomp}
\usepackage{stfloats}
\usepackage{url}
\usepackage{verbatim}
\usepackage{graphicx}
\usepackage{cite}
\usepackage{booktabs}
\usepackage {multirow}
\usepackage{makecell}
\begin{document}

\title{Environment Modeling for Service Robots \\ From a Task Execution Perspective}

\author{\IEEEauthorblockN{Ying~Zhang,~\IEEEmembership{Senior Member,~IEEE,}
                          Guohui~Tian,~\IEEEmembership{Member,~IEEE,}
                          Cui-Hua Zhang,~\IEEEmembership{Member,~IEEE,} \\
                          Changchun Hua,~\IEEEmembership{Fellow,~IEEE,} Weili Ding,
                          and~Choon Ki Ahn,~\IEEEmembership{Senior Member,~IEEE}
                          \vspace{-1.5em}
                          }

\thanks{This work was supported in part by the National Natural Science Foundation of China under Grant No. 62203378, 62203377, 62073279, in part by the Hebei Natural Science Foundation under Grant No. F2024203036, F2024203115, F2022203097, in part by the S\&T Program of Hebei under Grants 236Z2002G, 236Z1603G, in part by the Science Research Project of Hebei Education Department under Grant BJK2024195, and in part by the National Research Foundation of Korea (NRF) grant funded by the Korea government (Ministry of Science and ICT) under Grant NRF-2020R1A2C1005449. \emph{(Corresponding author: Cui-Hua Zhang and Choon Ki Ahn.)}}

\thanks{Y. Zhang, C. Zhang, C. Hua, and W. Ding are with the School of Electrical Engineering, and Hebei Key Laboratory of Intelligent Rehabilitation and Neuromodulation, Yanshan University, Qinhuangdao, 066004, China. (e-mail: yzhang@ysu.edu.cn; cuihuazhang@ysu.edu.cn; cch@ysu.edu.cn; weiye51@ysu.edu.cn;).

G. Tian is with the School of Control Science and Engineering, Shandong University, Jinan, 250061, China (e-mail: g.h.tian@sdu.edu.cn).

C. K. Ahn is with the School of Electrical Engineering, Korea University, Seoul 136-701, South Korea (e-mail: hironaka@korea.ac.kr).

{\color{blue}DOI: 10.1109/JAS.2025.125168}}
}

\markboth{This article has been accepted for publication in a future issue of the \emph{\textbf{IEEE/CAA Journal of Automatica Sinica}}.}%
{Shell \MakeLowercase{\textit{et al.}}: Environment Modeling for Service Robots From a Task Execution Perspective}


\maketitle

\begin{abstract}
Service robots are increasingly entering the home to provide domestic tasks for residents. However, when working in an open, dynamic, and unstructured home environment, service robots still face challenges such as low intelligence for task execution and poor long-term autonomy (LTA), which has limited their deployment. As the basis of robotic task execution, environment modeling has attracted significant attention. This integrates core technologies such as environment perception, understanding, and representation to accurately recognize environmental information. This paper presents a comprehensive survey of environmental modeling from a new task-execution-oriented perspective. In particular, guided by the requirements of robots in performing domestic service tasks in the home environment, we systematically review the progress that has been made in task-execution-oriented environmental modeling in four respects: 1) localization, 2) navigation, 3) manipulation, and 4) LTA. Current challenges are discussed, and potential research opportunities are also highlighted.
\end{abstract}

\begin{IEEEkeywords}
Environment modeling, service robot, mapping, task execution, long-term autonomy.
\end{IEEEkeywords}

\section{Introduction}
Due to recent advances in artificial intelligence (AI) and robotics \cite{tong2024advancements, kunze2018artificial, shen2024journey}, an increasing number of service robots have been developed for use in human-robot shared environments to provide services, such as 
object transportation tasks in the factory \cite{sirintuna2024object}, delivery tasks in the home \cite{zhang2020exploring}, and rescue tasks in the unknown scenario \cite{gao2021boundary}. 
One of the most fundamental factors affecting the intelligence of robotic task execution is the degree to which it is aware of the environment \cite{liu2024cognitive, liu2022service}. 
For example, in order to accomplish the tasks described above, a robot needs to accurately perceive and model the scene of its workspace as well as specific objects. In particular, home service robots have received significant attention, as they can be used to address some of the social problems caused by population aging and increase the overall quality of life. However, the intelligence of robots performing domestic service tasks in the home environment remains low. Unlike other human environments, the home environment is more complex \cite{zhang2020exploring, kunze2018artificial}
in the following respects:
\vspace{-0.07em}
\begin{itemize}
  \item \emph{It is open}.
  The home environment is not always delineated by physical partitions. 
      Many houses also have a personalized and diversified layout, increasing the sense of openness.
  \item \emph{It is unstructured}.
  The objects in the home environment are various and randomly positioned, which increases the diversity of the environmental characteristics and structures.
  \item \emph{It is dynamic}.
  Most objects in the home environment are movable. Their position, and posture often change during task execution. 
   The presence of people increases the dynamic characteristics of the environment further.
\end{itemize}

\begin{figure}[!t]
  \centering
  \includegraphics[width=2.7in]{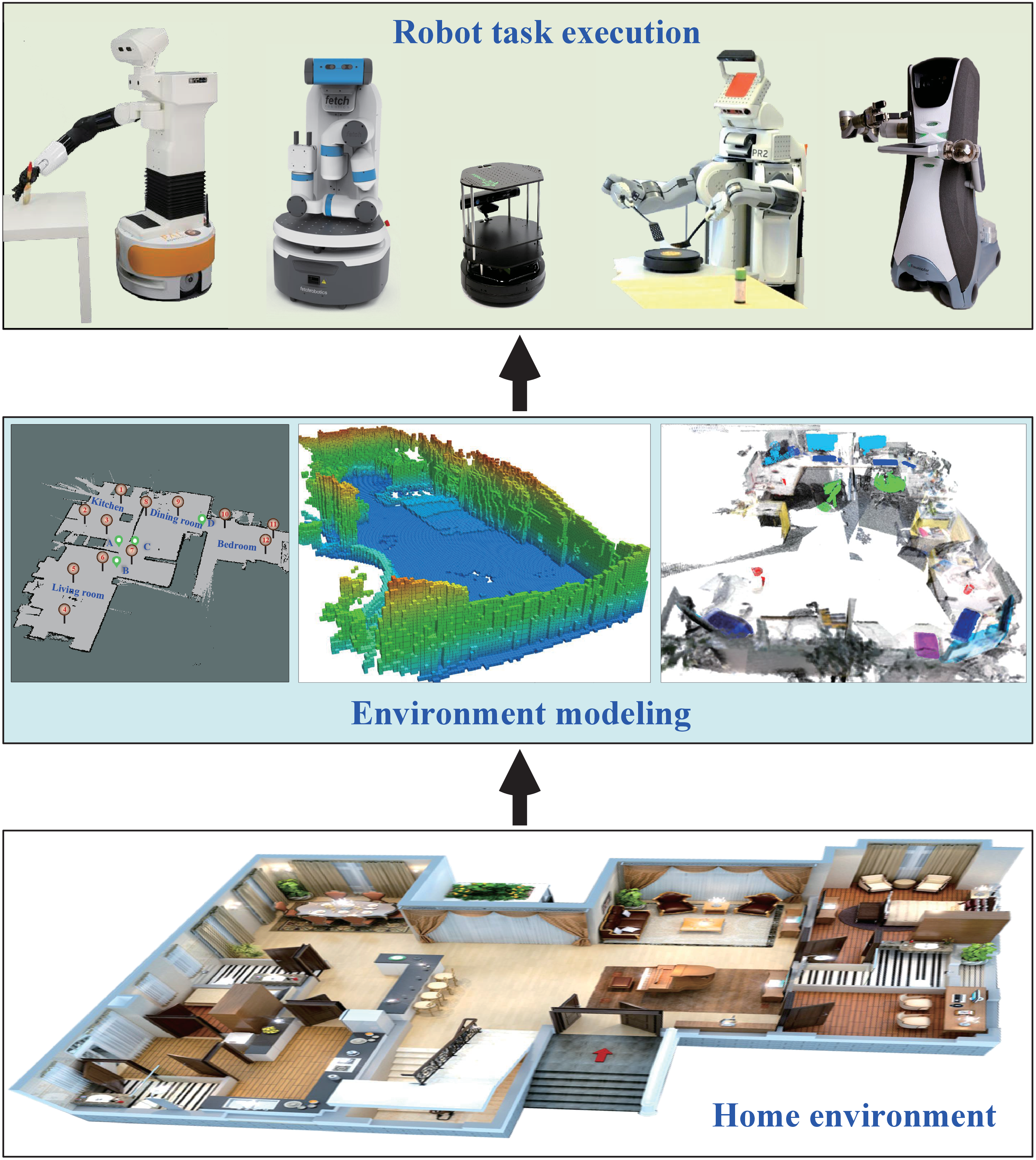}
  \caption{Basic process for robot task execution. A prerequisite for home service robots to perform domestic tasks is to model the environment as a 2-D, 3-D, semantic, or hybrid map to support robot localization, navigation, and manipulation \cite{wang2017autonomous, sunderhauf2017meaningful, zhang2021building}.}\label{flow}
\end{figure}
\vspace{-0.07em}
These factors increase the requirements for service robots in terms of accurately perceiving and understanding the home environment. A direct method to facilitate this is to develop a robot-oriented environmental model (or environmental map) and then issue the robot the environmental knowledge required for task execution \cite{wang2017autonomous, sunderhauf2017meaningful, zhang2021building}. 
As illustrated in Fig. \ref{flow}, the geometric layout, scenes, objects, and corresponding semantic information of the home environment can be modeled as a two-dimensional (2-D), three-dimensional (3-D), semantic, or hybrid map to support robot localization, navigation, and object manipulation. Past research has produced highly efficient, and high-quality environmental models that enable robots to complete expected service tasks in specific scenarios (e.g., such as relying on known objects, or deployed markers) \cite{ersen2017cognition, zhang2021safe}. However, the development of service robots that can perform everyday household chores in the open, unstructured, and dynamic natural home environment still has limitations.
Imagine a home scene where a service robot is tasked with delivering a box of milk from the kitchen to the living room. To complete the task, the robot usually models the environment efficiently in advance, and then obtains accurate location and semantic information of milk, kitchen, living room, etc. based on the environment model to perform the task. Besides, the presence of residents often changes the location of the target object, and such change is typically not perceived by the robot in time, which makes it impossible for the robot to efficiently perform the task at hand based on the constructed environment model, posing challenges particularly for the long-term effectiveness of robotic task execution.
In order for service robots to become full-time, reliable companions, they must be able to perform domestic service tasks efficiently, safely, and in a stable manner within the complex home environment over long periods.
In this context, constructing a task-execution-oriented environmental model by considering the relationships between robot tasks, scenes, and objects is a promising solution. Due to the open, unstructured, and dynamic nature of scenes, robotic task-execution-oriented home environment modeling remains a challenging research topic. To this end, this paper investigates environment modeling that supports four aspects: localization, navigation, manipulation, and long-term autonomy (LTA), guided by the needs of service robots for performing domestic service tasks.

Numerous surveys and review articles related to environmental modeling have been reported. In 2006, Durrant-Whyte and Bailey \cite{durrant2006simultaneous, bailey2006simultaneous} surveyed simultaneous localization and mapping (SLAM), summarizing its history, formulation, and solutions and analyzing the key challenges facing SLAM at the time. In addition, Santos \emph{et al.} \cite{santos2013evaluation} conducted a study of several laser-based 2-D mapping techniques, while Huang and Dissanayake \cite{huang2016critique} provided a critical review of the fundamental properties of the SLAM problem. Cadena \emph{et al.} \cite{cadena2016past} reviewed past work on the construction of environment models, covering multiple topics such as robustness and scalability in long-term mapping, metric and semantic mapping, and active SLAM. Furthermore, Kazerouni \emph{et al.} \cite{kazerouni2022survey} provided a survey of vision schemes for mapping. Younes \emph{et al.} \cite{younes2017keyframe} presented a survey of keyframe-based monocular mapping. Saputra \emph{et al.} \cite{saputra2018visual} introduced a review of visual SLAM in dynamic environments from visual localization and 3-D reconstruction techniques, while Pu \emph{et al.} \cite{pu2023visual} provided a review of visual SLAM based on semantic segmentation and deep learning, with a focus on the impact of dynamic objects on feature extraction and mapping accuracy. Further, Placed \emph{et al.} \cite{placed2023survey} investigated the current status and progress of active SLAM technology from a historical perspective and discussed related work on multi-robot active SLAM. Moreover, 
Xia \emph{et al.} \cite{xia2020survey} surveyed the research progress of semantic SLAM in three areas: perception, robustness, and accuracy. Crespo \emph{et al.} \cite{crespo2020semantic} summarized semantic environment modeling for robot navigation, including concepts, methods, and techniques for semantic information.
Han \emph{et al.} \cite{han2021semantic} proposed an overview of the research methods for the semantic mapping of indoor scenes, in which semantic mapping, spatial mapping, semantic acquisition, and map representation are discussed. Wang \emph{et al.} \cite{wang2024survey} reviewed visual SLAM in dynamic environments from a geometric to semantic perspective.
Additionally, many other related studies (e.g., \cite{eyvazpour2023hardware, al2024review, zhang2024air}) have been presented.

These previous studies have focused primarily on the history, formulation, and structure of environmental modeling problems and offered solutions to improve the accuracy and efficiency of modeling. Although the present paper also discusses some of these approaches, unlike previous reviews, it focuses on environment modeling for home service robots and comprehensively surveys the most commonly used techniques from a new perspective of task-execution-oriented environment modeling, which is guided by the requirements of home service robots designed for domestic tasks. In particular, state-of-the-art environment modeling methods for robot localization, navigation, manipulation, and LTA are discussed. The main contributions of this paper are as follows:
\begin{enumerate}
  \item A comprehensive survey of environmental modeling methods for home service robots is presented from a new perspective. To the best of our knowledge, this paper is the first to summarize robot task-execution-oriented environment modeling.
  \item Task-execution-oriented environment modeling methods are classified based on the requirements of home service robots designed to perform domestic tasks, and state-of-the-art modeling methods are discussed in detail.
  \item Current challenges facing environment modeling for home service robots are highlighted and future research directions are suggested.
\end{enumerate}

The remainder of this paper is organized as follows. Section \ref{section2} presents the features of robot task-execution-oriented environment modeling. In Sections \ref{section3} to \ref{section6}, environment modeling methods for robot localization, navigation, manipulation, and LTA are summarized, respectively. Finally, a conclusion and the outlook for future are provided in Section \ref{section7}.

\section{Features of robot task-execution-oriented environment modeling} \label{section2}
\begin{figure}[!t]
  \centering
  \includegraphics[width=2.6in]{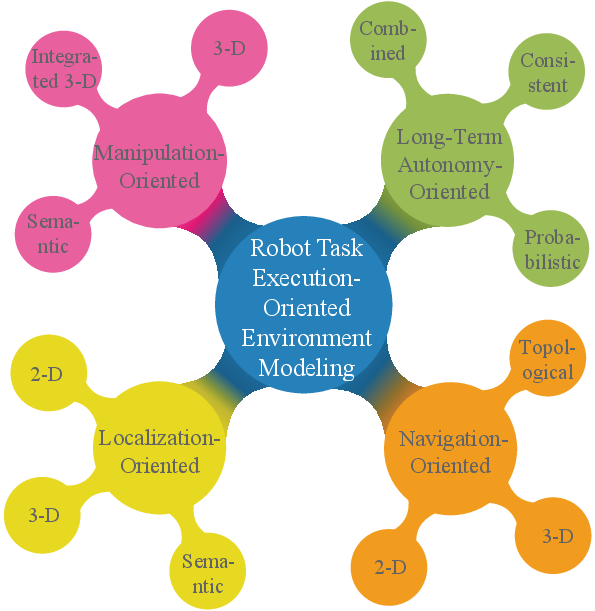}
  \caption{Four parts of robot task-execution-oriented environment modeling.}\label{components}
\end{figure}

This paper investigates the environment modeling problem for home service robot task execution. The purpose of this modeling is to allow a robot to operate safely and reliably in a complex home environment and to perform domestic service tasks over long periods. In order to achieve this goal, robots need to be able to identify their position accurately, navigate the house safely, manipulate objects effectively, and perform domestic tasks with LTA.
Therefore, robot task-execution-oriented environment modeling mainly includes: localization, navigation, manipulation, and LTA (refer to Fig. \ref{components}). To this end, we analyze modeling methods according to these four areas and provide research suggestions. As shown in Fig. \ref{components}, these methods are further subdivided based on the characteristics of the information involved.

\section{Robot localization-oriented modeling methods} \label{section3}
\begin{figure}[!t]
  \centering
  \includegraphics[width=3.3in]{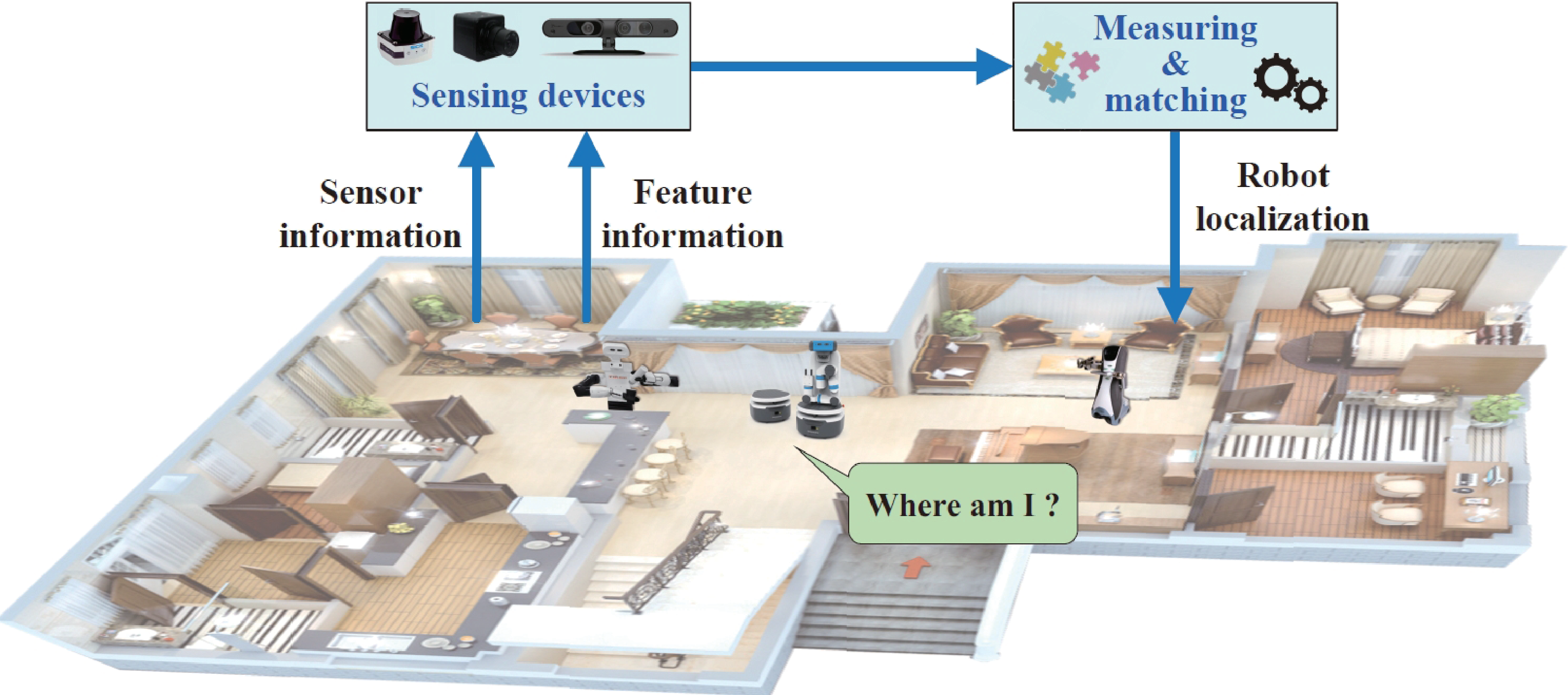}
  \caption{Typical process for robot localization.}\label{localization}
\end{figure}

Robot localization determines the position and posture of a robot within an environment and is the most basic foundation for task execution, refer to Fig. \ref{localization}. The solution is to collect and identify information from the surrounding environment using sensors installed on the robot itself, such as a laser rangefinder (LRF) or vision sensor, which is known as SLAM \cite{durrant2006simultaneous, bailey2006simultaneous}. SLAM has been the focus of a large volume of robotics research since it was first proposed \cite{smith1990estimating}, and a number of excellent research achievements have been reported to date.

\subsection{Two-dimensional models} \label{section31}
\begin{figure}[!t]
  \centering
  \includegraphics[width=3.4in]{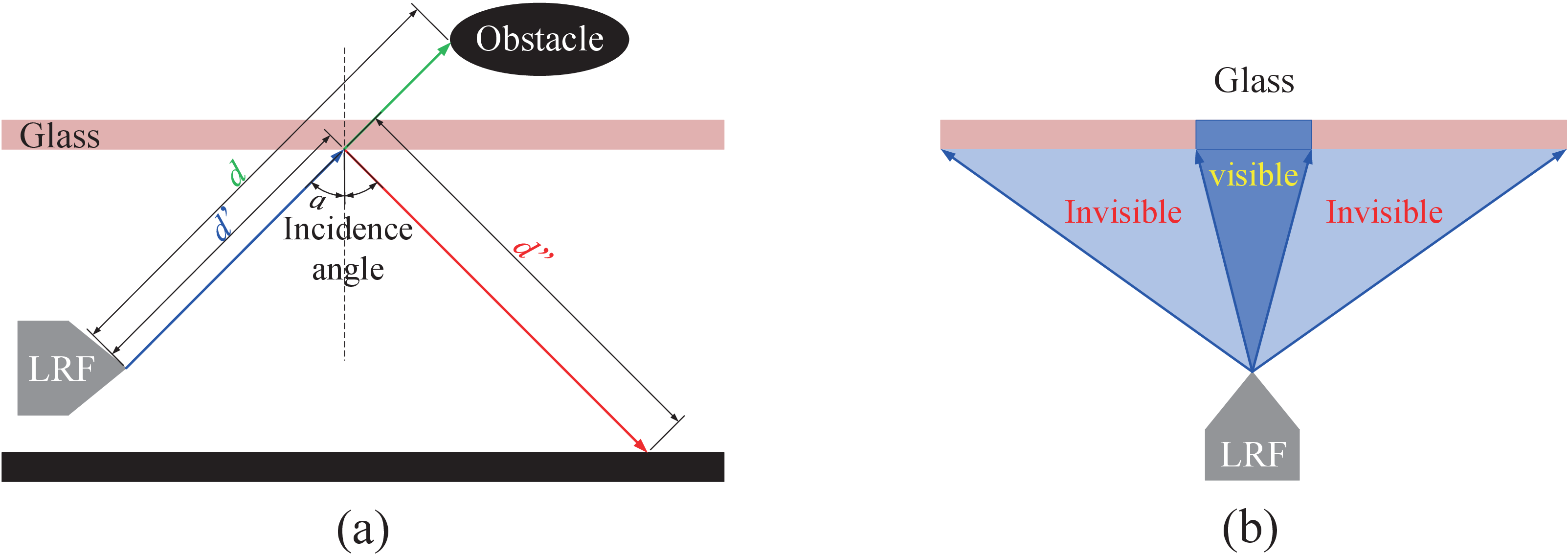}
  \caption{LRF measurement characteristics. (a) Three possible range measurements (i.e., $d$, $d'$ and $d''$) for glass walls \cite{kim2016localization}. (b) Glass can only be effectively detected within a limited angle close to its surface normal \cite{jiang2020online}.}\label{LRF}
\end{figure}

During task execution, robot localization usually depends on the construction of an environment map, with localization accuracy often directly affected by the map accuracy. Therefore, it is particularly important to build an accurate map. Classic methods such as 
KartoSLAM \cite{konolige2010efficient}, and Gmapping \cite{grisetti2007improved} have been widely used to create indoor 2-D maps. For example, Yilmaz \emph{et al.} \cite{yilmaz2019self} constructed a 2-D grid map based on LRF using Gmapping to achieve accurate indoor localization for a robot. Similarly, Wang \emph{et al}. \cite{wang2024exploration} relied on a 2-D occupancy grid map obtained from LRF for robot and target localization. To cope with the dynamic changes in the home environment, Krajn{\'\i}k \emph{et al.} \cite{krajnik2016persistent} proposed a time-dependent mapping system with LRF observations for persistent localization. The system uses the spatio-temporal 2-D grid to learn long-term environmental dynamics.
Similarly, Banerjee \emph{et al.} \cite{banerjee2023lifelong} presented a life-long 2-D mapping approach for robot localization in dynamic environments. However, because LRF is optical in nature, the reflection characteristics of an environment containing glass or other transparent objects can strongly affect the localization performance of a robot \cite{kim2016localization,jiang2020online} (refer to Fig. \ref{LRF}).

In order to improve the localization accuracy of a robot in the environment with glass walls, Kim and Chung \cite{kim2016localization} designed a scanning matching algorithm based on laser reflection characteristics and then proposed a LRF-based localization strategy. Awais \cite{awais2009improved} also used a probabilistic method to model the behavior of an LRF with a glass surface for robot localization. Similarly, Jiang \emph{et al.} \cite{jiang2020online} proposed a glass confidence map based on an LRF that could correctly characterize the occupancy state and probability of glass/non-glass, thus allowing the robot to take advantage of an environment model for its accurate localization within an environment containing glass. However, because LRFs can only scan the horizontal plane of an environment, they offer less information about the environment characteristics \cite{zhang2021effective}, meaning that they are not as useful for localization in geometric and/or dynamic environments. This, in turn, may lead to problems such as unsafe task execution.

As a response to this, Lee \emph{et al.} \cite{lee2019monocular} proposed a 2-D map construction scheme based on a monocular camera. In this solution, the robot position is estimated by matching the observation of the robot at different positions.
By converting RGB-D images into a mosaic, Da Silva \emph{et al.} \cite{da2020monocular} combined the descriptive ability of convolutional neural networks (CNNs) to estimate the robot position in an indoor environment.
Nevertheless, visual sensors are susceptible to environmental texture information and lighting conditions, which makes purely vision-based solutions tend to be unreliable. Benefiting from the advantages of an LRF, such as its high measurement accuracy and stable operation, 2-D environment modeling and robot localization based on LRFs remain the most common methods within the home environment.

\subsection{Three-dimensional models} \label{section32}
Compared with LRFs, visual sensors can obtain more features from the environment \cite{zhang2021effective}. For this reason, 3-D environment modeling for robot localization based on vision sensors have received significant attention. For example, Fu \emph{et al.} \cite{fu2019robust} designed a 3-D environment modeling system for indoor robots using RGB-D cameras, in which a line-based inliers refinement (LBIR) pose solver is used to process the correspondence between points and lines.
The localization of the robot is then realized by matching the feature information within the environment. Using a vision camera, Mur-Artal \emph{et al.} \cite{mur2015orb} combined features from accelerated segment tests (FAST) detection and binary robust independent elementary feature (BRIEF) descriptors for feature matching to achieve robot localization in a 3-D environment.
However, the performance of vision-based robot localization is strongly affected by a range of external factors, including environmental texture features and illumination.

To address these challenges, Maniscalco \emph{et al.} \cite{maniscalco2017robust} proposed a multi-sensor-based robot localization scheme. This scheme uses LRF to construct a 2-D map and then adopts Adaptive Monte Carlo Localization (AMCL)-based algorithms for robot localization based on multi-sensor data. By considering spatial occupancy and elevation information, Rico \emph{et al}. \cite{rico2024open} proposed a robot localization framework based on a 3-D environment model using RGB-D camera, 2-D and 3-D laser sensors, where the environment model is represented by the grid maps and Octomaps. Based on the 2-D grid map, Yu \emph{et al}. \cite{yu2023complex} used a 3-D laser sensor to conduct 3-D model of the ceiling for robot localization of the. In the method, geometric features on the ceiling are used as natural landmarks for robot localization. However, in a real home environment, the geometric features of the ceiling are similar, so these features are fragile for accurate robot localization. Moreover, Marder-Eppstein \emph{et al.} \cite{marder2010office} installed an LRF on the mobile base of a robot and a tilted platform at shoulder height to detect the 3-D environment for the robot localization. Zhen \emph{et al.} \cite{zhen2017robust} proposed a 3-D environment map construction method for robot localization based on a rotatable 2-D LRF. In this method, the LRF provides a global pose reference, an error state Kalman filter is then utilized to fuse the measurement data from the inertial measurement unit (IMU) for the robot localization in a 3-D environment. However, this type of 3-D measurement scheme based on a 2-D LRF and the control of the rotating motor is not very friendly to the localization of the robot.

\subsection{Semantic models} \label{section33}
For a service robot working in the home environment, simply being able to receive structural information from the environment is not sufficient. It is also necessary to understand semantic information related to the scene for robot localization. Research on environment modeling for semantic localization of the robot is thus useful for the performance of high-level complex tasks. For example, Premebida \emph{et al.} \cite{premebida2017dynamic} studied the semantic localization of robots in indoor scenes. Specifically, a dynamic Bayesian mixture model based on time is used to infer the semantic location of the robot based on the geometric characteristics of the environment perceived by the 2-D LRF.
On the basis of a 2-D grid map, Rosa \emph{et al.} \cite{rosa2019semantic} adopted two-way interaction between a robot and a user to allow the robot to infer its position in the human-robot environment. However, this method relies on user interaction information, which can affect the autonomy of the robot.

In addition, Rottmann \emph{et al.} \cite{rottmann2005semantic} proposed an environment modeling method to support robot semantic localization. This method uses the AdaBoost algorithm to extract environmental feature information from laser measurement data and visual observation data. A hidden Markov model (HMM) is then utilized to divide the indoor environment into rooms with different functions (such as the kitchen, office, etc.) to allow the robot to perform semantic localization. Similarly, Balaska \emph{et al.} \cite{balaska2020unsupervised} proposed an environment model construction framework to assist in the semantic localization of a robot by combining a 2-D grid map and the visual features of a scene.
Besides, Song \emph{et al.} \cite{Song2024RobotLOCA} presented a robot localization solution in dynamic indoor environments with similar layouts based on object and scene semantics. The solution first determines the area to which the robot belongs by comparing the semantic scene graph, and then uses a 2-D map built based on 2-D LRF to achieve precise localization of the robot. But the LTA of the robot is not considered.
To achieve lifelong updates of semantic localizations in dynamic environments, Narayana \emph{et al.} \cite{narayana2020lifelong} used the spatial transfer of semantics to update the 2-D map.
However, these methods can only achieve localization at the room level, and localization performance is significantly affected by the observation angle of the robot.

In order to enable a robot to determine its location within a complex indoor environment, Gomez \emph{et al.} \cite{gomez2023localization} introduced a 3-D semantic map model construction method combining metric and semantic information for robot localization, which associates depth and semantic data from RGB-D sensors with semantically enhanced truncated signed distance field (TSDF) at pixel level to estimate the robot pose. To achieve accurate localization, this method has certain requirements for perceptual robustness to movable objects and changes in object appearance.
Theoretically, compared with a 2-D environment model, a robot using a 3-D environment model should be able to achieve higher localization accuracy because more features can be obtained from the 3-D model. However, in practical applications, building a 3-D environment model not only leads to poor quality but also requires greater computational power. The low-quality environment model will in turn affect the localization accuracy of the robot.

\subsection{Comparative analysis} \label{section34}

\begin{table*}[!t]
\centering
\renewcommand{\arraystretch}{1.3}
  \centering
  \caption{Summary of representative works discussed above on robot localization}\label{table1}
  \scriptsize
  \renewcommand\arraystretch{1.2}
  \begin{tabular}{p{0.8cm}p{1.9cm}p{4cm}p{9.5cm}}
        \toprule
        \makecell[c]{Model} & \makecell[c]{Method} & \makecell[c]{Merit} & \makecell[c]{Demerit} \\
        \midrule
        \multirow{3}*{2-D} & \multirow{2}*{LRF \cite{do2018rish}, \cite{kim2016localization}} & \multirow{2}*{High measurement accuracy, stability} & Describe one horizontal plane of the environment; Not suitable for environments containing glass and other transparent objects or geometric environments with few features \\
         & Monocular \cite{lee2019monocular} & Obtain more environmental features & Susceptible to environmental texture information and lighting conditions \\
         \multirow{2}*{3-D} & Visual \cite{mur2015orb} & Match more environmental features & Strongly influenced by external factors such as texture features and illumination \\
         & IMU and LRF \cite{zhen2017robust} & Robust to motion, low drift & Not very friendly to robot localization with rotatable LRF \\
         \multirow{3}*{Semantic} & Interaction \cite{rosa2019semantic} & Infer the semantic position of the robot & Over-reliance on interactive information, affecting the autonomy of the robot \\
         & 2-D \cite{rottmann2005semantic} & Semantic positioning at the room level & Unable to locate in the room; Affect positioning performance by the robot's observation angle \\
         & 3-D \cite{taira2019right} & High semantic localization & Greater computational demands; 3-D semantic positioning is affected by the model quality \\
        \bottomrule
    \end{tabular}
\end{table*}

In summary, useful results have been achieved from robot localization modeling, as shown in Table \ref{table1}. LRFs are widely used in robot localization with 2-D, 3-D, and semantic environment models due to their advantageous measurement stability and accuracy. From the perspective of robot localization performance, the efficiency and accuracy of environment modeling, 2-D grid maps based on LRFs play a vital role. However, because an LRF can only observe environmental features along a horizontal plane, LRF-based 2-D maps cannot accurately describe the geometric structure of the environment, which may lead to problems such as unsafe task execution. In addition, Visual sensors can be used to obtain the 3-D characteristics of an environment, thus overcoming the limitations of LRFs to a certain extent \cite{zhang2021effective}. Despite this, the use of 3-D maps based on vision sensors poses greater challenges in terms of localization accuracy and computational efficiency. In this context, the use of LRFs and vision sensors to build a 2-D environment model that can accurately describe the environment is a promising avenue of research. This would not only allow robots to achieve accurate localization but also allow them to perform domestic tasks safely and efficiently.

\section{Robot navigation-oriented modeling methods} \label{section4}
\begin{figure}[!t]
  \centering
  \includegraphics[width=3.3in]{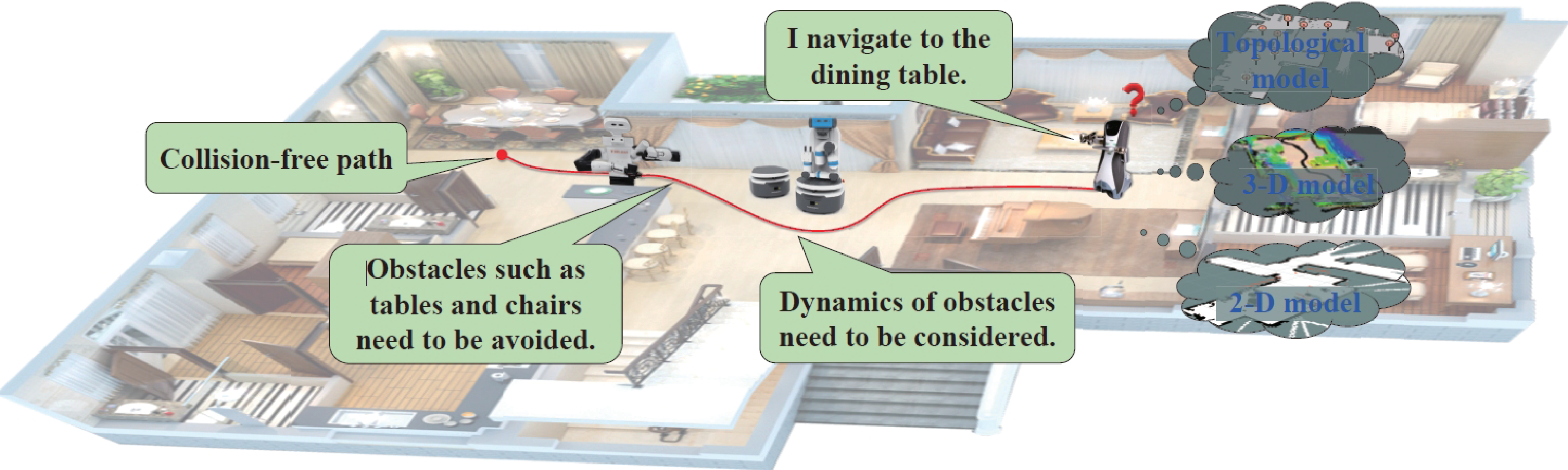}
  \caption{Basic process for robot navigation.}\label{navigation}
\end{figure}

Safe robot navigation is vital for service robots to perform domestic tasks in the home environment. Its purpose is to enable a robot to autonomously move from the current position to the target position without a collision during task execution (Fig. \ref{navigation}). The localization accuracy of the robot thus also plays a key role in the navigation process.

\subsection{Two-dimensional models} \label{section41}
\begin{figure}[!t]
  \centering
  \includegraphics[width=3in]{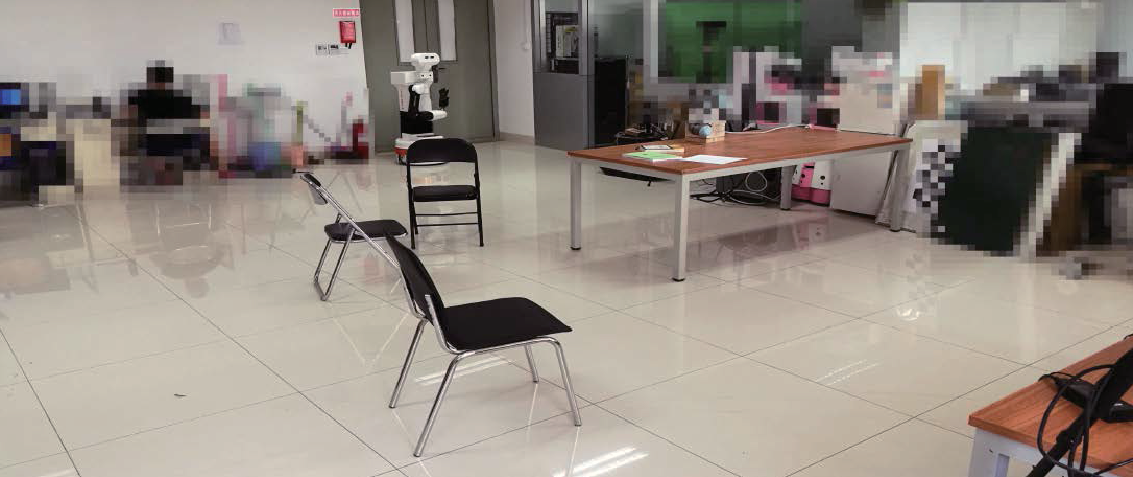}
  \caption{An example image of challenging scenarios \cite{zhang2021effective}. When an LRF-based robot navigates in this scenario, it may be able to avoid the legs of the table but then collide with the tabletop.}\label{Scenario}
\end{figure}

In order to support the accurate localization and effective navigation of robots, 2-D grid maps have been widely used. 
Biswas \emph{et al.} \cite{biswas2013localization} investigated the long-term navigation of robots in indoor environments using LRF-based 2-D grid maps. Besides, Or{\v{s}}uli{\'c} \emph{et al.} \cite{orvsulic2019efficient} proposed a 2-D occupancy grid map creation system based on boundary detection to support robot navigation. By considering user preferences, Zhang \emph{et al.} \cite{zhang2021user} presented a novel 2-D grid map construction method that integrates virtual areas to support human-aware robot navigation in the home environment. In addition, Song \emph{et al.} \cite{song2017safe} designed a guidance strategy based on an LRF to ensure the safe passage of robots within an indoor environment. Yani \emph{et al.} \cite{yani2023investigation} utilized LRF measurements to predict invisible obstacles on 2-D grid maps to improve indoor navigation performance. 
As mentioned in Section \ref{section3}, LRFs can only detect one horizontal plane of the environment. Therefore, when a robot uses an LRF-based 2-D grid map to navigate, its safety will be threatened in some challenging scenarios (see Fig. \ref{Scenario}). As an example, when a robot enters the dining room, it may be able to avoid the legs of the table but then collide with the tabletop. This is because the LRF cannot observe the tabletop information.

In response to this problem, numerous researchers have developed an alternative strategy using visual sensors. Compared with LRFs, visual sensors are less expensive and can generate dense depth information \cite{zhang2023crosslevel}. In particular, the use of 3-D measurement systems such as monocular and stereo vision to detect obstacles that are not fully visible to an LRF has been investigated. For example, in order to improve the navigation safety of a robot in an indoor corridor environment, Padhy \emph{et al.} \cite{padhy2019monocular} developed an autonomous navigation model using monocular vision to detect obstacle boundaries. Mora \emph{et al.} \cite{mora2024leveraging} suggested using a 3-D laser to generate a 2-D occupied grid map. The map contains the complete geometry of the object to allow a robot to safely navigate the environment. Based on homoplasy, Murray \emph{et al.} \cite{murray1997stereo} designed a robot guidance method to ensure the safe operation of the robot. In addition, Zug \emph{et al.} \cite{zug2012laser} suggested separating horizontal data from the data generated by a Kinect to build a 2-D map for robot task execution. However, this separation strategy ignores some necessary obstacle information, which affects the safety of robot navigation.

To ensure the stable navigation of a robot, Zhang \emph{et al.} \cite{zhang2021effective} proposed a 2-D grid map construction scheme for safe robot navigation based on laser-visual fusion, which converts visual observations into LRF-like data, and then builds the 2-D map representing spatial obstacles based on LRF-like and LRF data to improve robot navigation performance.
Qi \emph{et al.} \cite{qi2020object} combined a laser and vision camera to generate object-oriented 2-D map for robot navigation tasks in the domestic environment.
Luo \emph{et al.} \cite{luo2014multisensor} also introduced a method based on the combination of a stereo camera and an LRF to 
assist the robot in constructing an indoor map and in autonomous navigation. Baltzakis \emph{et al.} \cite{baltzakis2003fusion} proposed a laser and vision data fusion method to understand the structural information of a scene and applied it to robot navigation tasks in a corridor environment. However, the combination of laser and visual data relies on current observations of potential obstacles and has a weak correlation between the current and previous observation data. Moreover, it is difficult to determine whether visual or laser observation data most accurately describes obstacles in a particular situation. If this choice is wrong, a large error will occur in robot localization, which will result in an inaccurate map and affect the safety of robot navigation.

\subsection{Three-dimensional models} \label{section42}
A possible feasible solution is to adopt a 3-D environment model that can fully characterize obstacle information. Lau \emph{et al.} \cite{lau2013efficient} proposed a grid-based 3-D representation method for solving navigation tasks in dynamic environments.
Gregorio \emph{et al.} \cite{de2017skimap} proposed SkiMap, a multi-level environment model framework for robot navigation. SkiMap includes a 3-D voxel grid map, a 2.5-D height map, and a 2-D occupancy grid map, and supports continuous updates during robot navigation. Sathyamoorthy \emph{et al.} \cite{sathyamoorthy2024mim} presented a multi-layer intensity map (MIM) construction method for robot autonomous navigation. The MIM consists of multiple stacked 2-D grid maps for 3-D modeling of the environment, and each grid map is generated by the intensity of the reflection point cloud at a certain height interval.
Silva \emph{et al}. \cite{silva2023online} designed a robot navigation framework that takes user behavior into account by constructing a 3-D Octomap, where a sampling-based path planning method is used for robot navigation. With the advantages of Neural Radiance Fields (NeRF) in 3-D scene representation and high-fidelity rendering \cite{tosi2024nerfs}, Kim \emph{et al}. \cite{kim2024rnr} presented a renderable neural radiance (RNR) mapping framework for robot localization and navigation in 3-D environments, in which, the particle filtering and information loss reduction solution are used to improve mapping quality and navigation performance. Katragadda \emph{et al}. \cite{katragadda2024nerf} proposed a robot navigation system based on NeRF maps, which fuses \emph{a priori} NeRF maps in a tightly coupled manner. However, these methods requires a certain accuracy from the a priori 3-D map. Moreover, for home environments, the environment often changes due to the frequent movement of task-related objects. In this context, the images acquired by the robot and the a priori 3-D map will become mismatched. The 2-D maps typically characterize the static features of the environment, such as layout and structure. For this reason, due to the dynamic nature of the objects, the navigation performance of robots based on 3-D maps in dynamic and open home scenarios is generally lower than that based on 2-D maps.

In this regard, Wang \emph{et al.} \cite{wang2017autonomous} proposed a map construction method for robot safe navigation in an uneven and unstructured indoor environment based on a 2-D LRF and an RGB-D camera. The authors first represent the environment in the form of a 3-D grid OctoMap and then project the OctoMap onto a multi-layer 2-D occupancy grid map to generate a map that can be safely navigated. Similarly, Maier \emph{et al.} \cite{maier2012real} adopted a similar approach to map a 3-D environment model represented using depth cameras onto a 2-D grid map to better support robot autonomous navigation. Furthermore, Liu \emph{et al.} \cite{liu2024toward} designed a NeRF-based accessibility map generation method for long-term robot navigation tasks. In this method, an off-line trained NeRF is first used to model the 3-D spatial environment, and then the 2-D accessibility map is generated to represent space occupancy from the top-down view by assuming the equivalence between density and space occupancy.

The methods above all first require the construction of a 3-D model of the environment to fully describe the obstacle information and then project the constructed 3-D model onto a 2-D grid map that is more suited to robot navigation. This strategy can ensure the safety and navigation performance of a robot, but constructing a 3-D environment model requires significant computing power \cite{eyvazpour2023hardware}. In addition, the robot still needs to construct and map the environment model in real time during the navigation process. Obviously, this also requires greater computing power and navigation efficiency. Moreover, the semantic information for a scene and objects was not taken into account in these previous studies.

\subsection{Topological models} \label{section43}
In order to improve the navigation efficiency of robots, many researchers have modeled the environment as a topological map, with topological nodes representing different locations within the environment. 
Muravyev \emph{et al.} \cite{muravyev2023evaluation} investigated the deployment of topological maps in order to support the autonomous and efficient navigation of mobile robots. Blochliger \emph{et al.} \cite{blochliger2018topomap} designed the Topomap architecture, which extracts free space from sparse visual features to create a topological map representation of the environment, thus allowing for robot's navigation tasks. Gomez \emph{et al.} \cite{gomez2020hybrid} proposed a global topological representation method for large-scale indoor scenes, in which each room is modeled as a sub-map of the topological map for robot navigation. In order to support robot semantic navigation in structured indoor environments, Bavle \emph{et al.} \cite{bavle2022situational} used higher-level features to model rooms and corridors as topological nodes. In this case, a robot can efficiently navigate at the room level with the help of topological information. Zheng and Pronobis \cite{zheng2019pixels} introduced the TopoNets to create a semantic topological map of a large-scale environment, with the rooms classified according to the geometric characteristics of the scene, and robot navigation is conducted in combination with the underlying 2-D grid map. Though these methods allow topological navigation between rooms, the efficient navigation of a robot within a room was not investigated.

In order to effectively improve the planning and navigation capabilities of robots in indoor environments, Song \emph{et al}. \cite{song2024fht} built a topological map containing main nodes and support nodes based on visual features. Although the robot can realize indoor topological navigation based on this topological map, this method creates too many redundant nodes that are not related to the task. In addition, Fraundorfer \emph{et al.} \cite{fraundorfer2007topological} proposed a topological environment model based on scene images. Based on this model, the robot uses an image search and matching to achieve the navigation task.
However, this method requires a large number of images to be collected, which makes image retrieval and maintenance difficult.
In order to generate a spatially meaningful topological map, Kim \emph{et al.} \cite{kim2023topological} used images in which objects can be detected as topological points to establish a topological semantic graph memory for robot navigation. 
Similarly, Delfin \emph{et al.} \cite{delfin2018humanoid} proposed a construction method for topological maps based on visual memory. This method uses a human-guided approach to capture key images offline. Robot localization and navigation are realized by searching for key images and matching them to the currently observed images. But this method requires significant manpower for the learning process and the coding, and topological maps based on key images cannot reliably support robot navigation in a dynamic environment, especially when the scene in the position of the key images changes.

\subsection{Comparative analysis} \label{section44}
An overview of the main methods discussed above is exposed in Table \ref{table2}.
In general, LRF-based 2-D environment models allow a robot to efficiently navigate a space but, in a complex home environment, the safety of the robot can be threatened. A 3-D environment model can describe scene information more comprehensively, but its construction has a higher computational burden for the robot, which also still faces the problem of low navigation efficiency based on this type of model. To improve the efficiency of robot navigation, modeling the environment as a topological structure is a common technical approach, such as the construction of a topological map based on scene clustering, movement distance, and visual memory. Although these strategies can improve the navigation efficiency of robots to a certain extent, the dynamics of the objects within the environment and the relevance of the robotic tasks have not been considered in the process of creating topological nodes, meaning that these topological maps cannot deal with a dynamic environment \cite{zhang2021building}. In addition, the autonomous execution of domestic tasks by robots requires efficient navigation in open and dynamic home scenarios. Therefore, under the premise of considering robotic tasks, establishing an environment model that is conducive to the efficient navigation of a robot is vital to improving its intelligence and promoting its use in the home environment.

\begin{table*}[!t]
\centering
\renewcommand{\arraystretch}{1.3}
  \centering
  \caption{Summary of representative works discussed above on robot navigation}\label{table2}
  \scriptsize
  \renewcommand\arraystretch{1.2}
  \begin{tabular}{p{0.8cm}p{2.2cm}p{6cm}p{7.4cm}}
        \toprule
        \makecell[c]{Model} & \makecell[c]{Method} & \makecell[c]{Merit} & \makecell[c]{Demerit} \\
        \midrule
        \multirow{3}*{2-D} & LRF \cite{zhang2021user} & High navigation efficiency; Support human-aware navigation & Navigation security at risk in challenging scenarios with spatial obstacles \\
         & Kinect \cite{zug2012laser} & Separate horizontal observation data to represent obstacles & Ignore necessary obstacle information, affecting navigation safety  \\
         & LRF and vision \cite{baltzakis2003fusion} & Extract accurate scene structure information & Weak data correlation; difficulty in selecting data describing obstacles  \\
         \multirow{2}*{3-D} & Monocular  \cite{katragadda2024nerf} & Support robot navigation in the given 3-D map & Need for accurate a priori maps; Require significant computing power \\
         & LRF and RGB-D \cite{maier2012real} & Better support robot autonomous navigation & Build 3-D models beforehand; Require significant computing power \\
         \multirow{2}*{Topological } & Depth camera \cite{gomez2020hybrid} & Support semantic understanding and room-level navigation & Effective navigation of the robot in the room is not available \\
         &Digital camera \cite{fraundorfer2007topological} & Highly scalable, allowing operation in large environments & Need to collect a large number of images; Difficult to retrieve and maintain \\
        \bottomrule
    \end{tabular}
\end{table*}

\section{Robot manipulation-oriented modeling methods} \label{section5}
\begin{figure}[!t]
  \centering
  \includegraphics[width=2.5in]{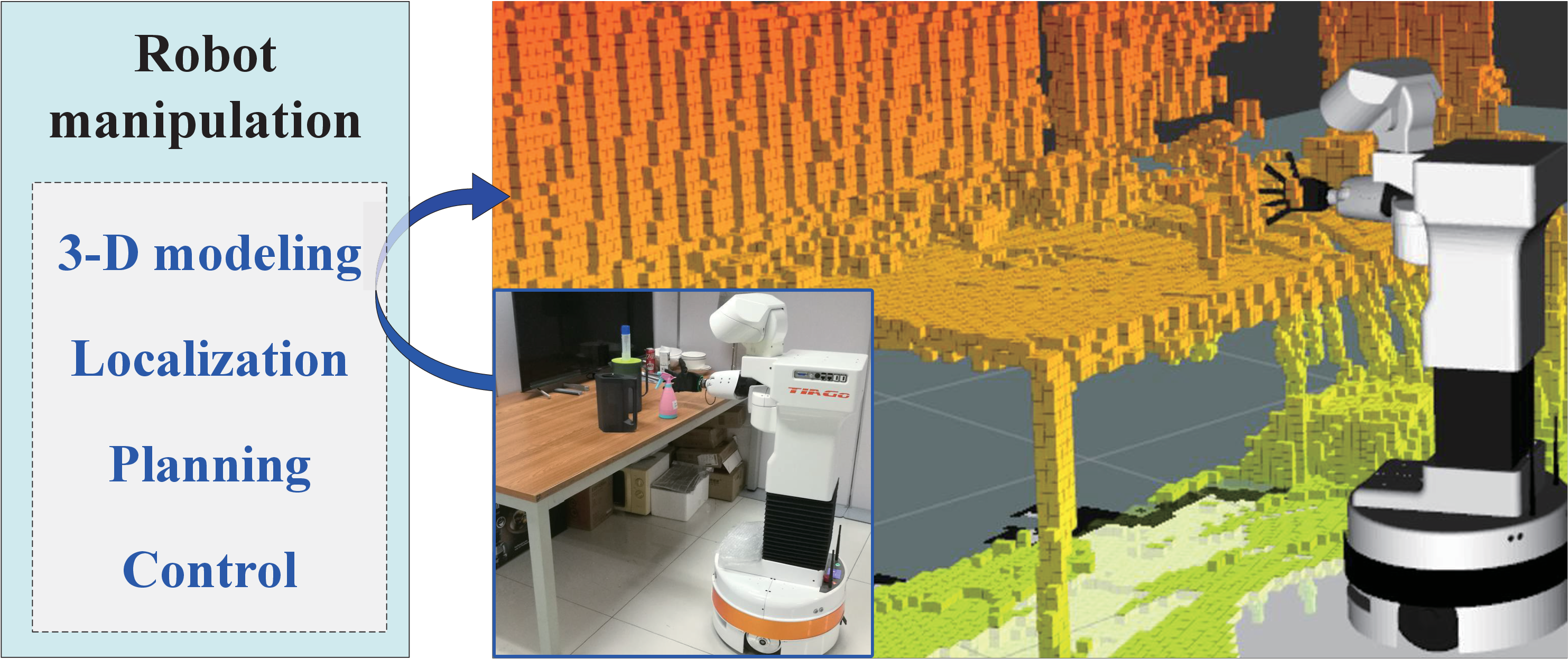}
  \caption{Illustration of 3-D environment modeling for robot manipulation \cite{zhang2021safe}.}\label{manipulation}
\end{figure}

In addition to localization and navigation, service robots deployed in the home environment need to be able to effectively perform object manipulation to complete domestic tasks. Because this manipulation is carried out in a 3-D environment, 3-D environment modeling is required (Fig. \ref{manipulation}). 

\subsection{Three-dimensional models} \label{section51}
For manipulation tasks, 3-D environment modeling is a key prerequisite for home service robots. This provides a volumetric representation of the 3-D space within the environment, and it has attracted considerable research attention \cite{zhang2021safe}. In order to efficiently represent a 3-D environment, Hornung \emph{et al.} \cite{hornung2013octomap} proposed an octree-based 3-D environment modeling architecture, OctoMap, which uses the probabilistic method to estimate the state of the 3-D environment and represents the environment in compact form as occupied, free, or unknown. 
Gervet \emph{et al}. \cite{gervet2023act3d} introduced a 3-D action map of high spatial resolution construction method for multi-task robotic manipulation by representing the workspace as continuous a 3-D feature field. In order to allow robots to perform object manipulation tasks in cluttered or occluded environments, Bimbo \emph{et al.} \cite{bimbo2022force} proposed a 3-D environment occupancy modeling method that uses physical interactions to simultaneously construct objects and environments. To enable a robot to perform manipulation tasks in an unknown environment, Murooka \emph{et al.} \cite{murooka2016planning} introduced a contact sensor model and reported the construction and updating of a 3-D occupancy map based on contact information. However, this method requires the robot to be equipped with special contact sensors and the ability to perform groping behavior.

In addition, Hara \emph{et al.} \cite{hara2020moving} proposed a representation scheme based on a surface mesh map. In this approach, dynamic objects are removed, while static objects are reserved for task execution. Ran \emph{et al.} \cite{ran2023object} combined the object semantic and geometric features to propose a 3-D environment model that can provide robots with the semantic labels, object scales and poses needed to perform manipulation tasks.
Qiu \emph{et al}. \cite{qiu2024learning} used an RGB-D camera to scan the scene including the target object and constructed a 3-D representation based on the generalizable feature field (GeFF). With the help of the 3-D map and the target object, the robot performs mobile manipulation tasks.
Similarly, in order to allow robots to perform effective manipulation tasks, Monica \emph{et al.} \cite{monica2020point} deployed an octree-based framework for motion planning. Terasawa \emph{et al.} \cite{terasawa20203d} used an octree-based representation method to construct a 3-D occupancy grid and used it for robot collision detection and object grasping. By learning environment-aware affordance, Wu \emph{et al.} \cite{wu2023learning} made use of point-level representations for 3-D environment modeling to guide the robot to perform object manipulation tasks in occlusion scenarios.

\begin{figure}[!t]
  \centering
  \includegraphics[width=3.5in]{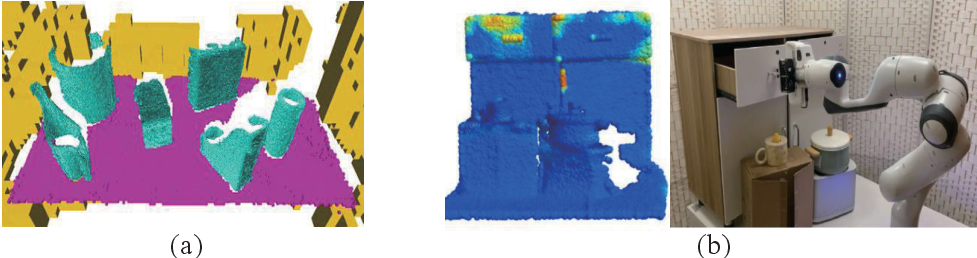}
  \caption{Examples of 3-D modeling of the environment supporting robot manipulation \cite{hornung2013octomap, wu2023learning}}\label{3Dmap}
\end{figure}

The methods above have established 3-D environment models to support robot manipulation tasks (see Fig. \ref{3Dmap}). These methods construct objects as part of the environmental model. In this context, the robot can efficiently perform housekeeping tasks in a static home environment using the constructed model, but it is challenging to do so in a dynamically changing environment \cite{ersen2017cognition}.
Although these methods can deal with changes in the environment by characterizing the operating scenes in real time, home service robots still need to handle an open environment. In practical applications, robots are often required to perform the manipulation of unknown objects \cite{zhang2019efficient}. In this case, robots using the above methods require 3-D modeling of the operating environment to be implemented again to maintain the consistency between the model and the scene. Obviously, this restricts the intelligence and adaptability of the robot when performing domestic service tasks.

\subsection{Integrated three-dimensional models} \label{section52}
In addition to the efficiency and accuracy of 3-D model representations, the construction of an integrated 3-D environment model that can describe object information is also particularly important in terms of providing guidance for service robots to perform object manipulation. Wu \emph{et al.} \cite{wu2023object} proposed an object-oriented integrated environment modeling framework for robot everyday manipulation that provides 3-D point clouds, object map and topological map for robots to perform grasping tasks and with object semantics, object poses, and other information related to the manipulated object itself.
Based on 3-D environment model, Lu \emph{et al.} \cite{lu2023ovir} presented a 3-D object instance modeling method for robot manipulation by multi-view fusion and smoothing, which allows appropriate grasping positions for robot manipulation to be identified. 
S{\"u}nderhauf \emph{et al.} \cite{sunderhauf2017meaningful} combined SLAM, object detection, instance-level segmentation, data association, object model updating, and other technologies to propose an object-oriented integrated 3-D environment model that provides instance-level object models and geometric point cloud information for task execution. However, although these solutions increase the convenience for robots in performing domestic tasks within the home environment, they are computer resource-intensive due to the requirement for an accurate full 3-D environment model to be constructed in advance to support robot manipulation. 

In order to reduce the computational costs, Megalingam \emph{et al.} \cite{megalingam20232d} proposed an effective 2-D - 3-D hybrid mapping method that allows 3-D perception and 2-D localization, path planning, and autonomous navigation using LRF and Kinect sensor.
Based on a 2-D occupancy grid, Martins \emph{et al.} \cite{martins2020extending} used a 3-D point cloud to model 3-D environment information, with the 2-D occupancy grid used for rapid planning and 3-D point cloud representation used for safe navigation and manipulation tasks. Similarly, Ruiz-Sarmiento \emph{et al.} \cite{ruiz2017building} constructed a point cloud model of 3-D scenes using RGB-D images on the basis of a 2-D occupancy grid map to support robot manipulation tasks. In addition, to address the low efficiency of robot navigation based on 3-D environment models, Gregorio and Stefano \cite{de2017skimap} proposed a multi-level full 3-D environment mapping framework in which a 2-D occupancy grid map can be obtained for robot navigation.
In like manner, Wang \emph{et al.} \cite{wang2020stable} presented a scheme to model the environment using a lightweight navigation map. Their solution down-projects a 3-D environment map into a 2-D description form and then generates a traversable map for the safe navigation of the robot. Based on homoplasy, Biswas and Veloso \cite{biswas2012depth} projected 3-D observation data into a 2-D model for robot navigation, while Hornung \emph{et al.} \cite{hornung2012navigation} mapped a 3-D environment model based on an octree onto a 2-D occupancy map and proposed an integrated method using 2-D and 3-D environment representation to address the low efficiency of 3-D path planning during mobile manipulation tasks.

Although the above methods offer efficient path planning and navigation, they still require a full 3-D representation model of a scene to be constructed in advance \cite{zhang2021safe}. In addition, using these methods, a robot needs to continuously perform real-time 3-D modeling of the working environment during the manipulation task to handle changes to the scene. This does not reduce the computational burden on the robot and represents an inflexible approach. At present, research in this area is still in its exploratory stages.

\subsection{Semantic models} \label{section53}
The semantic understanding of environmental information and adaptability to a dynamic home environment are key factors in improving the intelligence of service robots for manipulation tasks. Therefore, it is necessary to build an environment model that supports environmental semantic representation and dynamic reasoning, which can enhance a robot's in-depth understanding of the environment. Zhang \emph{et al.} \cite{zhang2021safe} proposed a semantic environment model for safe and efficient robot manipulation by integrating a 2-D grid map, real-time local 3-D representation, and an object ontology model.
Schmalstieg \emph{et al.} \cite{schmalstieg2023learning} used a semantic map as a central memory component for robot mobile manipulation. The map is extended to a local panoptic segmentation map with object instance labels to support object manipulation. G{\"u}nther \emph{et al.} \cite{gunther2017model} presented a hybrid semantic modeling method for the indoor environment. This method matches the object geometric model with the point cloud for verification to enrich the object model required by the robot to perform manipulation tasks.
Zhang \emph{et al.} \cite{zhang2023hierarchical} also took advantage of ontology technology to create an environment model that provides the robot with corresponding knowledge for serving tasks through query and reasoning.
Using environmental maps, ontology technology to describe environmental knowledge has become an effective means of constructing and conceiving home environmental knowledge, but there is still a lack of in-depth research that addresses all of the logical relationships and the accurate reasoning between objects.

Riazuelo \emph{et al.} \cite{riazuelo2015roboearth} introduced the RoboEarth semantic mapping system, which uses ontology to encode the concepts of maps and objects and the relationships between them. Their constructed environment model includes SLAM maps to provide scene geometry knowledge and object locations. To complete daily manipulation tasks,
Kwak \emph{et al.} \cite{kwak2022semantic} utilized a constructed semantic model of robotic manipulation as an information source to provide relevant knowledge for service robots. Similarly, Bi \emph{et al.} \cite{bi2023object} proposed a task-oriented environment model in combination with a semantic map, in which a constructed knowledge base is used as an information source to provide service robots with relevant knowledge such as objects and their positional relationships to complete daily manipulation tasks.
Ruiz-Sarmiento \emph{et al.} \cite{ruiz2017building} expressed diversified environmental knowledge and used a conditional random field (CRF) model and environmental context to deal with object uncertainty. 
Based on contextual clues, Rogers \emph{et al.} \cite{rogers2012conditional} employed a CRF model to achieve joint reasoning between rooms and objects. In this case, the robot is able to infer the location and range of the room through the modeled object-room compatibility and adjacency dependency.

Though the above methods allow for the reasoning of modeled environmental knowledge, the relationship between different objects and between objects and rooms will change due to the presence of users and robot actions. This will result in inaccurate relationships within the created environment model. However, this has not been considered in past research. At the same time, service robots deployed a dynamic and open home environment require LTA to perform domestic tasks, which has higher requirements for comprehensive and accurate modeling and reasoning.

\subsection{Comparative analysis} \label{section54}

\begin{table*}[!t]
\centering
\renewcommand{\arraystretch}{1.3}
  \centering
  \caption{Summary of representative works discussed above on robot manipulation}\label{table3}
  \scriptsize
  \renewcommand\arraystretch{1.2}
  \begin{tabular}{p{0.8cm}p{2.1cm}p{5.2cm}p{8.3cm}}
        \toprule
        \makecell[c]{Model} & \makecell[c]{Method} & \makecell[c]{Merit} & \makecell[c]{Demerit} \\
        \midrule
        \multirow{3}*{3-D} & RGB-D \cite{zhang2021safe} & Real-time mapping; Collision-free motion planning & With the help of auxiliary markers; Weak adaptation to strong dynamic scenarios \\
         & Contact sensor \cite{murooka2016planning} & Achieve groping behavior; Safe trial motion & Need for special contact sensors; Low efficiency of robot manipulation \\
         & 3-D camera \cite{monica2020point} & Grasp planning and part-based semantic grasping & Inability to manipulate unknown objects; Poor adaptability to dynamic environment  \\
         \multirow{2}*{Integrated} & 3-D object \cite{wu2023object} & Provide objects' point clouds, semantics, etc. & Computer resource-intensive; Poor adaptability to dynamic environment \\
         & 3-D projection \cite{wang2020stable} & Offer effective path planning and navigation & High computational burden; Poor dynamic adaptability; Inflexible \\
         \multirow{2}*{Semantic} & Ontology \cite{zhang2021safe} & Object relationship model for robot manipulation & Lack of deeper description of environmental knowledge \\
         & CRF \cite{ruiz2017building} & Categorize the percepts of objects and rooms & Difficulty in modeling dynamic environments; Inaccurate reasoning \\
        \bottomrule
    \end{tabular}
\end{table*}

In Table \ref{table3}, we provide a brief overview of the representative works that have been described above.
Overall, the 3-D modeling of the environment is important for robot manipulation tasks. Therefore, most current solutions employ the full 3-D characterization of a robot's working environment. This approach models task-related objects as part of the environment, but there are certain limitations in dealing with the dynamics of the environment. In addition, the efficiency of robot planning and navigation based on 3-D environment models is low. For this reason, a large number of studies have transformed their 3-D environment models into 2-D form or directly constructed a joint 2-D and 3-D representation of the environment. However, these methods still require an a priori full 3-D model of the environment and, during the task execution, the robot also requires a 3-D representation of the scene in real time to deal with changes in the environment. This is not a flexible strategy, and the computational burden on the robot is also higher. In contrast, the utilization of ontology technology to describe the logical relationship between objects and scenes based on environmental maps is effective in dealing with the uncertainty of scenes. Although a large volume of research has been conducted on the representation of environmental knowledge, there is still a lack of research on the comprehensive and accurate representation of environmental knowledge and task-driven dynamic uncertain reasoning within a home environment.

\section{Robot LTA-oriented modeling methods} \label{section6}
\begin{figure}[!t]
  \centering
  \includegraphics[width=3.2in]{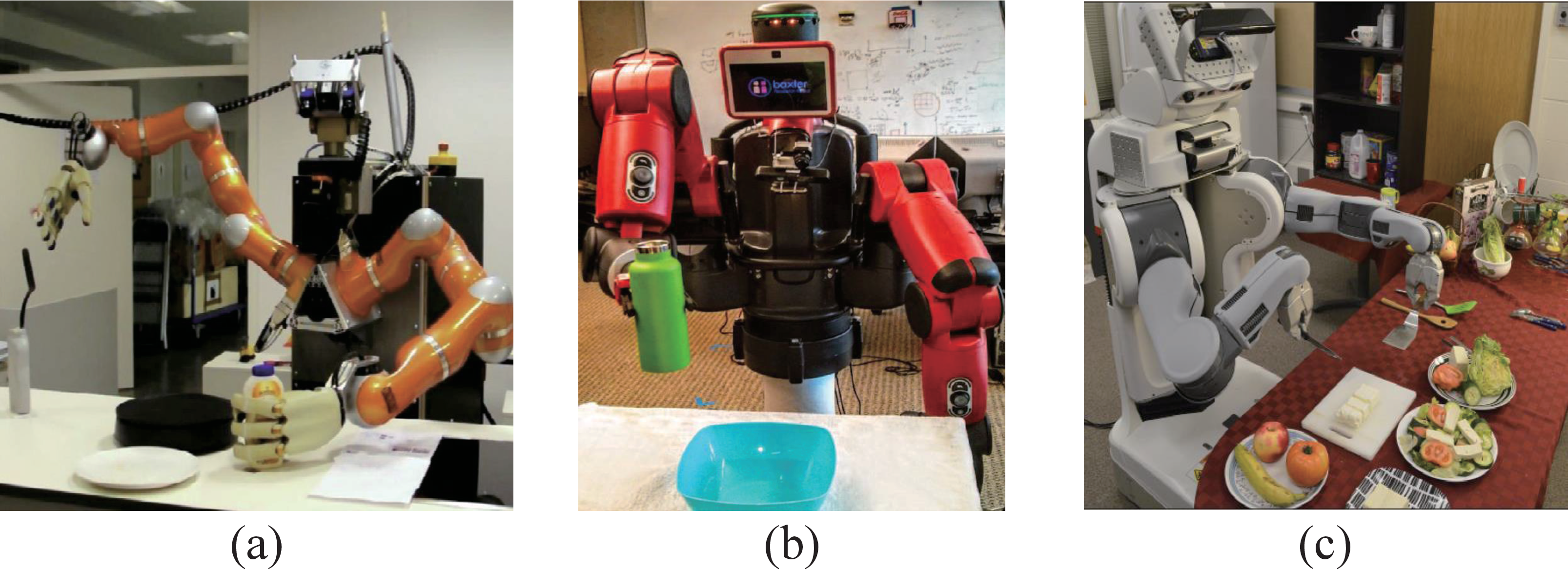}
  \caption{Examples of a service robot performing domestic tasks: (a) making pancakes based on simulation-based projections \cite{kunze2017envisioning}, (b) pouring water in bright lighting conditions \cite{schenck2018perceiving}, (c) preparing meals by perceiving the physical properties of objects in advance \cite{gemici2014learning}.}\label{domestictask}
\end{figure}

Based on localization, navigation, and manipulation, service robots deployed in a home environment are expected to perform domestic tasks with LTA (Fig. \ref{domestictask}). Many studies have reported excellent progress in achieving this goal.

\subsection{Combined models} \label{section61}
The use of a combined model of the environment to assist service robots in performing domestic tasks has been demonstrated experimentally. For example, Kunze \emph{et al.} \cite{kunze2017envisioning} constructed a combined environment model based on a physics engine to allow a robot to execute complex service tasks, such as making pancakes [see Fig. \ref{domestictask}(a)]. To realize such tasks, the authors deeply embedded the reasoning component in the robot control program, allowing task- and context-dependent decisions to be made and action parameters to be determined based on the underlying robot components.
Hauser \emph{et al.} \cite{hauser2011randomized} also proposed a task execution method for the service robot by fusing multi-modal motion planning in the constructed environment model, with artificial markers deployed to assist the robot in performing complex manipulation tasks. In order to enable robots to learn and perform services such as pouring water [see Fig. \ref{domestictask}(b)], Schenck \emph{et al.} \cite{schenck2018perceiving} introduced a liquid detection and tracking scheme based on deep CNNs based on the RGB-D and thermal cameras. Based on fiducial markers, Zhang \emph{et al.} \cite{zhang2021safe} proposed a combined model for robot manipulation that integrates a 2-D model, a 3-D model, and an object model. Inceoglu \emph{et al.} \cite{inceoglu2019continuous} proposed a combined environment modeling system based on vision and marker assistance to enable a service robot to handle everyday manipulation tasks and complete household chores.
In order to improve the adaptability of robot task execution, Cui \emph{et al}. \cite{cui2024active} developed a fusion modeling mechanism based on graph attention network to characterize task-related semantic features by combining scene knowledge, object knowledge and their positional relationships.

Many of the successful applications described above currently rely on some combination of known target objects, simplified static environments, defined artificial markers, or specific task controllers \cite{ersen2017cognition}. These dependencies and constraints also limit the LTA of robots in performing domestic tasks in the home environment. In order for service robots to operate effectively in the home environment over long periods, they must have the ability to perform tasks efficiently in an unconstrained environment. 
Notably, service robots cannot fully observe the whole home environment using their own sensors, thus increasing the difficulty of reliably performing everyday tasks based on an a priori environment model.

\subsection{Consistent models} \label{section62}
In order to improve the LTA of service robots performing domestic tasks, it is necessary to establish and maintain an environment model that is consistent with the robot's working environment. The most direct approach to building a consistent environment model is to create a mapping relationship for all of the entities within the environment one by one \cite{zhang2022semantic}. For instance, a top-down approach has been adopted to define manual symbolic knowledge to produce a consistent hand-crafted model \cite{thosar2021multi}. Similarly, Hertzberg \emph{et al}. \cite{hertzberg2002learning} used a hand-crafted method to define chronicles for robot learning. To improve the robot's ability to perform daily household chores,  Li \emph{et al.} \cite{li2024semantic} manually created knowledge models of objects, service tasks, constraints, and rules in an attempt to achieve consistent representation with real task scenarios. In a static environment, this can be considered a feasible method because the robot can perform tasks efficiently based on a consistent hand-crafted model. However, when the robot is in the open, dynamically changing family environment, relying on this method can lead to task failure, which further prevents robots from completing housekeeping tasks with LTA. And when the environment changes, a large amount of coded information needs to be updated manually to maintain consistency with the entities. This not only poses challenges to manpower and efficiency but also significantly limits the LTA of service robots. In addition, one-to-one mapping methods lead to modeling environment information that is relatively independent, meaning the robot will lack reasoning ability.

To update the mapping relationship in the created environment model, Coradeschi \emph{et al.} \cite{coradeschi2003introduction} introduced a mapping architecture known as anchoring. Based on this architecture, the robot can autonomously update the mapping relationship according to the observed object properties. Persson \emph{et al.} \cite{persson2020semantic} proposed an anchored matching function to improve the overall anchoring process. Elfring \emph{et al.} \cite{elfring2013semantic} used a 2-D blob visual detector to determine the shape, size, and color of an object and employed this to create and update the mapping relationship for the environment model. In order to enhance mapping performance, Shih \emph{et al.} \cite{shih2016exploiting} described an elegant form of anchoring that exploits the luminance and chrominance of objects. In order to facilitate the deployment of robots in the home environment, Anderson \emph{et al.} \cite{anderson2018vision} adopted the visual information for the mapping environment in a similar way. For dynamic scenes, Zhang \emph{et al.} \cite{zhang2022semantic} proposed an environment modeling solution based on semantic grounding with visual perception for robot long-term task execution. In the solution, static objects and rooms are grounded in an interactive way to maintain consistency with the workspace.

Despite these advances, a robot cannot fully observe its working environment due to its own limited perception ability. Therefore, if the robot wants to maintain a consistent mapping relationship within the established environment model, then it needs to frequently traverse the entire working environment to update this relationship \cite{zhang2022semantic}. In addition, the use of a service robot's vision system to create and update the mapping relationship is susceptible to the observation conditions, object placement positions, and sensor noise. These factors may cause incorrect mapping results, leading to task failure and preventing the robot from working autonomously in the home environment over long periods.

\subsection{Probabilistic models} \label{section63}
In order to enhance a robot's LTA for task execution, a number of studies have attempted to establish a probabilistic model of the relationship between objects and rooms on the basis of environmental maps to give robots reasoning capabilities and enhance environmental adaptability. For example, based on the semantic map, Chikhalikar \emph{et al.} \cite{chikhalikar2024semantic} designed an environmental model for object search by utilizing the probabilistic relationships between objects that reflect the spatial relationships in the home environment, allowing the robot to infer the location of target object for task execution.
Mantelli \emph{et al.} \cite{mantelli2022semantic} provided a probabilistic and semantic modeling approach to build the environment map so that a robot can decide which regions to visit first to quickly find the target object.
Aydemir \emph{et al.} \cite{aydemir2011plan} used semantic location classification to construct a hierarchical probabilistic model for the environment so that the robot can infer where the target object is most likely to be found. Similarly,
Murali \emph{et al.} \cite{murali20206} developed a target-driven probabilistic method to reason about the unseen object parts in cluttered environment.
Wang \emph{et al.} \cite{wang2018efficient} made use of the semantic distance between the target object and the room to determine the room where the target object is most likely to appear in order to perform the task.

Despite their promising results, the methods summarized above do not consider the impact of the current position of the robot on task-execution efficiency, meaning that the inferred position may not necessarily be the optimal position of the target object. In addition, although the possible position of the task-related object can be inferred based on a probabilistic model, it is still difficult for a service robot to quickly locate a target object in chaotic or dense home scenarios and perform the domestic task with LTA.

To this end, Park \emph{et al.} \cite{park2023zero} created a vision-based semantic grid environment model with the help of static landmarks (e.g., tables), and introduced an indirect method designed to improve the search efficiency of robots in a complex home environment by guiding the robot to explore areas where task-related objects may be present.
Zeng \emph{et al.} \cite{zeng2020semantic} used the probabilistic relationship between objects to represent the environment as a semantic linking map. On this basis, task-related objects can be quickly located with the help of intermediate objects to enhance the LTA of the robot.
Similarly, Zhang \emph{et al.} \cite{zhang2019efficient} adopted three important linguistic descriptions of spatial relationships between objects (i.e., \emph{in}, \emph{on}, and \emph{next to}.) to guide the robot to efficiently approach a target object.
Based on RGB-D images and semantics, Honerkamp \emph{et al}. \cite{honerkamp2024language} constructed an environment model including the semantic 3-D map, navigation Voronoi graph, and room scene graph. Then, based on this model, the target object was localized by using a large language model (LLM) for reasoning about the exploration area and interacting with the environment to complete daily housekeeping tasks.
To enable a robot to effectively visit an inferred area, Kostavelis \emph{et al.} \cite{kostavelis2017semantic} deployed a metric-topological mapping framework that integrated visual cues, in which the belief distribution about the visited location is added during the exploration process.
However, the nodes on this map are determined using specific distances. In this case, too many redundant nodes irrelevant to the task at hand are created in the environment model, which greatly reduces task efficiency. Additionally, in the above methods, the impact of task execution on the modeled object-room relationship is not considered. At the same time, the openness of the home environment (for example, users may move or change the position of an object without the robot's knowledge) can lead to inaccurate relationships within the built environment model \cite{zhang2022semantic}, which will strongly influence the effectiveness 
of task performance, thus limiting the LTA of the robot.

\subsection{Comparative analysis} \label{section64}

\begin{table*}[!t]
\centering
\renewcommand{\arraystretch}{1.3}
  \centering
  \caption{Summary of representative works discussed above on robot LTA}\label{table4}
  \scriptsize
  \renewcommand\arraystretch{1.2}
  \begin{tabular}{p{1cm}p{1.9cm}p{5.8cm}p{7.7cm}}
        \toprule
        \makecell[c]{Model} & \makecell[c]{Method} & \makecell[c]{Merit} & \makecell[c]{Demerit} \\
        \midrule
        \multirow{2}*{Combined} & Integrated \cite{zhang2021safe} & Task-oriented object inference and robot manipulation & Manually define the relationships of object model \\
         & Fusion \cite{cui2024active} & Provide spatial information, types and properties of objects & Manual coding; Poor dynamic adaptability \\
         \multirow{2}*{Consistent} & hand-crafted \cite{hertzberg2002learning} & Coding chronicle definitions to ground fact symbols & Difficulty in maintaining consistency; Low adaptability in dynamic scenarios \\
         & Anchoring \cite{coradeschi2003introduction} & Update mapping relationships in the environment model & Weak update performance; Difficulty in maintaining consistency \\
         \multirow{2}*{Probabilistic } & Relationship \cite{zhang2019efficient} & Guide the robot to the most likely object location & Reliance on handcrafting; Limit the LTA of robot task execution\\
         & Extraction \cite{chikhalikar2024semantic} & Use prior relationships to increase the finding probability & Coding prior relationships; Poor adaptability in chaotic or dense scenarios \\
        \bottomrule
    \end{tabular}
\end{table*}

In conclusion, a considerable amount of effort has been devoted to enable service robots to perform domestic tasks, as shown in Table \ref{table4}. And some successful cases of robots performing complex domestic service tasks (such as making pancakes) have been reported, but most of these involve specific service tasks in a specific environment. These constraints highlight the lack of robot adaptability to the environment. In practical applications, service robots need to perform everyday household chores autonomously over long periods in a naive (i.e., unconstrained) home environment. Establishing and maintaining an environment model that is consistent with a robot's working environment, which involves the creation of mapping relationships among the entities in the environment one by one, is a feasible strategy to achieve this goal. This approach allows a service robot to perform tasks efficiently in a static environment or when the target object is known. However, the service robot usually does not know the location of the target object for the actual task. Moreover, the entities in this mapping method are relatively independent, which means that the service robot ultimately lacks reasoning ability and leads to its poor adaptability to the environment. To overcome this problem, a common method is to construct a probabilistic environment model based on the positional relationship between the objects and the rooms, so that a service robot can infer the position of a target object when performing a domestic task. Nevertheless, most current research has focused on the one-time execution of a task and has not taken into account potential inaccuracy in the modeled relationship arising from successful task execution. Service robots deployed in the home are expected to be able to perform housekeeping tasks over a long period rather than completing individual tasks only once. In addition, the openness and dynamics of the home environment increase the challenge for robots to perform tasks with LTA. Therefore, the modeling of the home environment for efficient and long-term autonomous task execution is a problem that requires further research.

\section{Conclusion and outlook} \label{section7}
The establishment of environment models that assist robots in performing domestic tasks efficiently and autonomously over long periods is vital to facilitating the entry of service robots into homes and the provision of intelligent services.
In this vein, the present paper proposed a comprehensive survey of task-execution-oriented environment modeling for home service robots. Guided by the requirements for robots for domestic tasks, we introduced environmental modeling approaches for four areas: robot localization, navigation, manipulation, and LTA (a summary of the methodology and evaluation can be found in the supplementary material). The issues and challenges were discussed to identify potential future research directions. The specific findings are summarized below.

First, robot localization modeling methods, including 2-D, 3-D, and semantic approaches, were outlined. In general, localization schemes based on 2-D environment models have played a key role in the execution of robotic tasks. However, these have limitations in describing 3-D spatial information.Affordable vision sensors have become more widely available, allowing more environmental features to be captured and providing 3-D sensing, which has inspired researchers to employ visual information to build in the construction of 2-D environment models. 
In the future, the fusion of multiple sensors such as LRF and vision in the construction of 2-D environment models that can represent the 3-D spatial information of the environment to better support robot localization and safe operation will be a promising area of research. Meanwhile, the Internet of Things (IoT) \cite{keroglou2023survey} can also be integrated to improve the localization performance of service robots deployed in home environments for a long time.

Second, robot navigation models, including 2-D, 3-D, and topological models, were introduced, and the advantages and disadvantages of various construction methods were discussed. Of these strategies, topological models are currently a widely used solution. However, state-of-the-art topological models ignore the correlation of robotic tasks in the construction process, so they cannot fundamentally improve the execution of these tasks. In addition, the dynamic nature of the environment and the diversity of tasks also pose challenges to the robot's long-term and efficient navigation. Therefore, the construction of an appropriate environment model for robot navigation that can improve the efficiency of task execution is vital to enhancing the intelligence of home service robots. To this end, future work should develop an environment model conducive to robot navigation, such as an effective combination of 2-D, semantic, and topological information, by considering robotic tasks and environmental complexity. On this basis, it is also particularly important to use machine learning technology to enable the robot to learn from experience and optimize the navigation strategy \cite{levine2023learning}.

Third, it was shown that manipulation by home service robots requires 3-D modeling of the environment. In this context, 3-D, integrated 3-D, and semantic models have been proposed. In general, these modeling methods not only require full 3-D modeling of the environment in advance but also require real-time 3-D representation to address environmental changes during task execution. Although some success has been achieved in practice, manipulation-oriented environment modeling still faces a number of challenges. For example, for robot task execution, a full 3-D environment model contains many redundant scenes (e.g., 3-D modeling of navigation scenes does not improve the convenience for the robot in task execution), which increases computing and memory costs. Hence, for home service robots, a flexible environment modeling method that supports efficient robot manipulation requires further development in future work, such as integrating promising techniques, e.g., NeRF, 3-D Gaussian Splatting, learned representations for geometric fields, and resource-efficient optimization algorithms \cite{tosi2024nerfs, katare2023survey}, to investigate task-driven environment modeling solution.

Fourth, combined, consistent, and probabilistic environment modeling methods were presented for use as LTA-oriented environment models, and the challenges facing these state-of-the-art approaches were discussed. However, most of these methods still rely on a priori assumptions or require human intervention to enable service robots to autonomously perform specific domestic tasks or one-time tasks. For a service robot to be deployed in the complex home environment over a long period, it needs the ability to perform domestic tasks autonomously without (or with less) human intervention. And note that the diversity of domestic tasks, the dynamic and open nature of the home environment also increase the requirements for LTA-oriented environmental modeling. It is worth mentioning that the technological advances in vision foundation models and embodied AI are enhancing the environmental awareness of service robots \cite{firoozi2024foundation}. Thus, future research should investigate the integration of vision foundation models, embodied AI and life-long learning into environmental modeling to enhance the universality and applicability of LTA-oriented environmental models.

In general, research on task-execution-oriented environment modeling is the key to improving the execution capability and intelligence of service robots, enabling home service robots to be reliably employed in practical applications. Despite the progress that has been made to date, there is still a long way to go before service robots are widely used in domestic households. In particular, there are a number of challenges that need to be addressed and uncharted territory to be explored. It is notable that, although this paper focused on task-execution-oriented modeling for home service robots, there are many other applications that could benefit from improved modeling heuristics, such as social robots, and autonomous vehicles.

\bibliographystyle{IEEEtran}
\bibliography{mybibfile}

\begin{IEEEbiography}[{\includegraphics[width=1in,height=1.25in,clip,keepaspectratio]{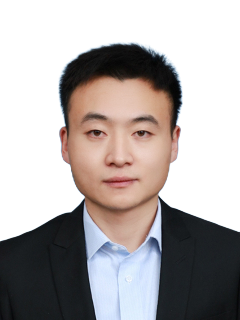}}]{Ying Zhang} (Senior Member, IEEE) received the Ph.D. degree in Control Theory and Control Engineering from Shandong University, Jinan, China, in 2021.

He is currently an Associate Professor with the School of Electrical Engineering, Yanshan University. He has authored or
coauthored more than 50 journal papers. He is an Early Career Associate Editor for \emph{Biomimetic Intelligence and Robotics}, \emph{Intelligence \& Robotics}, and \emph{Brain-X}. Dr. Zhang received Best Paper Award in CCIR 2018, the Youth Top Talent Project for Hebei Education Department in 2023, and the Excellent Youth Project for Hebei NSF in 2024. His current research interests include intelligent robot systems, visual perception, environment modeling, air-ground robotic collaboration.
\end{IEEEbiography}

\begin{IEEEbiography}[{\includegraphics[width=1in,height=1.25in,clip,keepaspectratio]{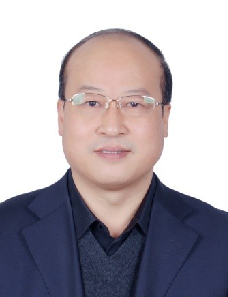}}]{Guohui Tian} (Member, IEEE) received the B.S. degree from the Department of Mathematics, Shandong University, Jinan, China, in 1990, the M.S. degree from the Department of Automation, Shandong University of Technology, in 1993, and the Ph.D. degree from the School of Automation, Northeastern University, Shenyang, China, in 1997.

From 1999 to 2001, he was a Post-Doctoral Researcher with the School of Mechanical Engineering, Shandong University. From 2003 to 2005, he was a Visiting Professor with the Graduate School of Engineering, Tokyo University, Tokyo, Japan. He was a Lecturer from 1997 to 1998 and an Associate Professor from 1998 to 2002 with Shandong University, where he is currently a Professor with the School of Control Science and Engineering. His current research interests include service robot, intelligent space, cloud robotics, and brain-inspired intelligent robotics.
\end{IEEEbiography}

\begin{IEEEbiography}[{\includegraphics[width=1in,height=1.25in,clip,keepaspectratio]{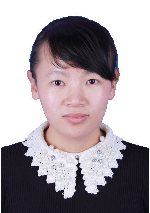}}]{Cui-Hua Zhang} (Member, IEEE)
received the M.S. degree in control theory and control engineering from the Institute of Automation, Qufu Normal University, Qufu, China, in 2017, and the Ph.D. degree in control theory and control engineering from Northeastern University, Shenyang, China, in 2021. 

She is currently an Associate Professor with the School of Electrical Engineering, Yanshan University, Qinghuangdao, China. Her current research interests include intelligent robot systems, nonlinear adaptive control and networked nonlinear systems.
\end{IEEEbiography}

\begin{IEEEbiography}[{\includegraphics[width=1in,height=1.25in,clip,keepaspectratio]{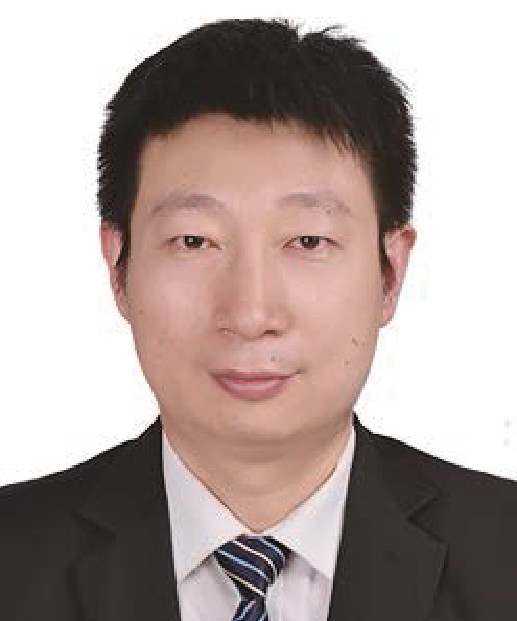}}]{Changchun Hua} (Fellow, IEEE/CAA) received the Ph.D. degree in Electrical Engineering from Yanshan University, Qinhuangdao, China, in 2005. From 2006 to 2007, he was a Research Fellow with the National University of Singapore, Singapore. From 2007 to 2009, he was with Carleton University, Ottawa, ON, Canada, funded by the Province of Ontario Ministry of Research and Innovation Program. From 2009 to 2010, he was with the University of Duisburg-Essen, Essen, Germany, funded by the Alexander von Humboldt Foundation.

He is currently a Full Professor with Yanshan University. He has been involved in more than 15 projects supported by the National Natural Science Foundation of China, the National Education Committee Foundation of China, and other important foundations. He is a Cheung Kong Scholars Programme Special Appointment Professor. He is an Associate Editor for the \textsc{IEEE Transactions on Cybernetics}, \emph{International Journal of Control, Automation and Systems}, and other journals. He is a highly cited Researcher (since 2014) selected by Elsevier. His research interests include nonlinear control systems, multiagent systems, teleoperation systems, and intelligent robot systems.
\end{IEEEbiography}

\begin{IEEEbiography}[{\includegraphics[width=1in,height=1.25in,clip,keepaspectratio]{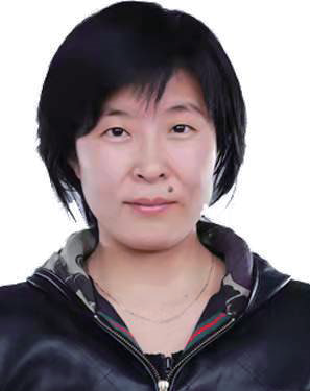}}]{Weili Ding} received the B.Sc. degree in automation from the Liaoning University of Technology, Jinzhou, China, in 2003, and the Ph.D. degree from the Shenyang Institute of Automation, Chinese Academy of Sciences, Beijing, China, in 2008.

She is currently a Full Professor with the Department of Automation, Yanshan University, Qinhuangdao, China. Her research interests include machine vision, image processing, and virtual reality.
\end{IEEEbiography}

\begin{IEEEbiography}[{\includegraphics[width=1in,height=1.25in,clip,keepaspectratio]{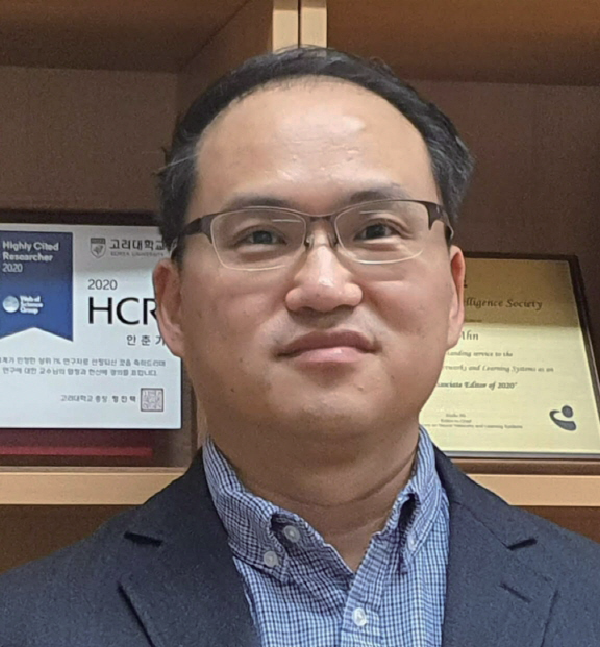}}]{Choon Ki Ahn} (Senior Member, IEEE) received the Ph.D. degree in the School of Electrical Engineering and Computer Science from Seoul National University, Seoul, Korea, in 2006. He is currently a \textit{Crimson Professor of Excellence} with the College of Engineering and a Full Professor with the School of Electrical Engineering, Korea University, Seoul, Korea. In 2017, he received the \textit{Presidential Young Scientist Award} from the President of South Korea. In 2020, 2021, and 2023, he received the Best (Outstanding) Associate Editor Award for \textsc{IEEE Transactions on Neural Networks and Learning Systems}; \textsc{IEEE Transactions on Systems, Man, and Cybernetics: Systems}; and \textsc{IEEE Transactions on Circuits and Systems I: Regular Papers}, respectively. In 2019-2024, he received the Korea University Research Excellence Award. In 2023, he was inducted into the Korea University Hall of Fame as the world's top researcher.

He has been a Senior Editor of \textsc{IEEE Transactions on Neural Networks and Learning Systems; IEEE Transactions on Systems, Man, and Cybernetics: Systems;} \textit{IEEE Systems Journal}. He has also been an Associate Editor of \textsc{IEEE Transactions on Fuzzy Systems; IEEE Transactions on Automation Science and Engineering; IEEE Transactions on Intelligent Transportation Systems; IEEE Transactions on Circuits and Systems I: Regular Papers}; \textit{IEEE Systems, Man, and Cybernetics Magazine,} and other flagship journals. He is the recipient of the 2019-2024 Highly Cited Researcher Award in Engineering by Clarivate Analytics (formerly, Thomson Reuters).
\end{IEEEbiography}

\end{document}